\PassOptionsToPackage{table}{xcolor} 
\documentclass[11pt]{article}

\usepackage[preprint]{acl}

\usepackage{times}
\usepackage{latexsym}

\usepackage[T1]{fontenc}

\usepackage[utf8]{inputenc}

\usepackage{microtype}

\usepackage{inconsolata}

\usepackage{graphicx}

\usepackage{amsmath}
\usepackage{amssymb}
\usepackage{amsfonts}
\usepackage{algorithm,algpseudocode}
\usepackage{adjustbox} 
\usepackage{diagbox}
\usepackage{multirow}
\usepackage{subfigure}
\usepackage{indentfirst}
\usepackage{setspace}
\usepackage{array}
\newcolumntype{C}[1]{>{\centering\arraybackslash}p{#1}}
\usepackage[inline]{enumitem}
\usepackage{adjustbox}
\usepackage{bbding}
\usepackage[table]{xcolor}
\usepackage[most]{tcolorbox}
\usepackage{courier} 
\usepackage{wrapfig}

\newtcolorbox{promptbox}{
    colback=gray!10,      
    colframe=gray!50,     
    boxrule=0.5pt,        
    arc=3pt,              
    left=6pt,
    right=6pt,
    top=6pt,
    bottom=6pt,
    fontupper=\ttfamily,  
}

\usepackage{hyperref}       
\usepackage{url}            
\usepackage{booktabs}       
\usepackage{amsfonts}       
\usepackage{nicefrac}       
\usepackage{xcolor}         
\usepackage[table]{xcolor}
\newcommand{\resultbest}[1]{\textbf{#1}}
\newcommand{\resultsecond}[1]{\underline{#1}}
\newcommand{\pivotmethod}[1]{\textbf{#1}}
\newcommand{\twoline}[2]{\shortstack{#1 \\ #2}}
\usepackage{caption}
\usepackage{graphicx}

\usepackage{pifont}
\newcommand{\xmark}{\ding{55}}%

\usepackage{textcomp}

\title{PIVOT: Bridging Planning and Execution in LLM Agents via Trajectory Refinement}

\author{%
  Tuo Zhang \\
  \texttt{tuozhang@amazon.com} \\
   \And
  Alin-Ionut Popa \\
  \texttt{popaaln@amazon.com} \\
  \AND
  Yan Xu \\
  \texttt{yanxuml@amazon.com} \\
  \And
  Rui Song \\
  \texttt{ruisong@amazon.com} \\
  \And
  Dimitrios Dimitriadis \\
  \texttt{dbdim@amazon.com} \\
}

\begin{document}
\maketitle

\begin{abstract}
When a long-horizon Large Language Model (LLM) agent fails, the decisive error is often made long before failure surfaces. Existing methods either critique the final output and replan globally, breaking parts that were already correct, or map failures onto a fixed error taxonomy that cannot express task-specific repairs. 
We explore whether an agent can instead locate the earliest execution-grounded causal error in its trajectory and repair only the affected suffix. To address it, we propose PIVOT, which couples planning and execution through four stages: PLAN states an explicit plan before acting; INSPECT checks execution evidence against the plan and localizes the earliest unsupported step; EVOLVE rewrites the trajectory from that step while preserving the validated prefix; and VERIFY checks the final output against all task constraints.
Empirical evaluations on DeepPlanning and GAIA demonstrate state-of-the-art performance: PIVOT establishes a strong upper bound up to $\sim$94\% relative improvement in constraint satisfaction, showing that the core trajectory-refinement mechanism remains effective without external supervision. PIVOT also remains computationally efficient, requiring up to 3--5$\times$ fewer tokens than competing refinement methods.

\end{abstract}

\section{Introduction}
\label{sec:intro}

\begin{figure*}[t]
  \centering
  \includegraphics[width=\linewidth]{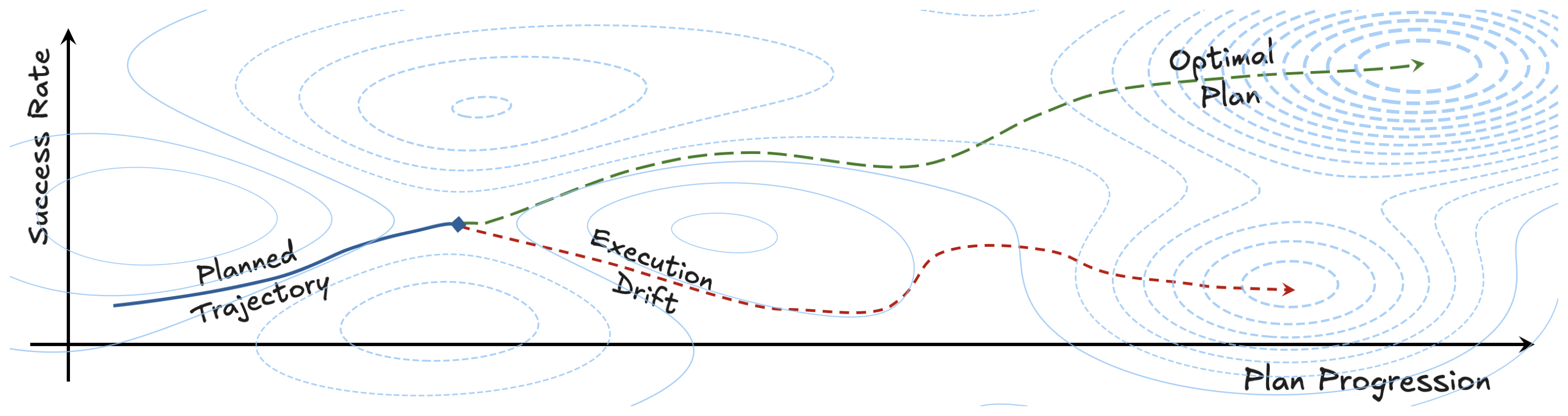}
  \caption{LLM-generated plans (blue) often appear valid but diverge during execution (red) due to infeasible steps, incorrect state assumptions, or constraint violations. These discrepancies compound over long horizons, leading to undesirable or suboptimal outcomes that surface only at the end. Aligning planning with execution to reach an optimal trajectory (green) is therefore a core challenge in reliable agentic systems.}
  \label{fig:teaser}
\end{figure*}

Long-horizon LLM agents make dozens of interdependent decisions, such as tool queries, entity selections, and schedule commitments, but receive feedback only on the final output~\cite{yao2023react}. This feedback could identify which constraint is violated at the end of the trajectory, but could not point out which decision caused the violation, which typically lies at an earlier step. As shown in Figure~\ref{fig:teaser}, as execution first diverges from the plan, the error propagates through every decision that often render the remainder of the plan unrecoverable~\cite{zhu2025llm}. This raises a central question for reliable long-horizon agents: can an agent use execution evidence to identify where its plan breaks down, both to prevent errors from propagating and to repair them from their origin?

Existing agent-improvement methods only partially address this challenge because they lack a trajectory-level optimization view. Diagnostic and attribution
methods~\cite{zhu2025llm,cemri2025multi,zhang2025agent,zhang2025agentracer,deshpande2025trail} identify failures post hoc, but map them onto fixed error categories and do not repair the trajectory; iterative refinement methods~\cite{madaan2023selfrefine,shinn2023reflexion,Gou+24,bohnet2025enhancing} critique the final output and revise it as a whole, discarding parts that were already correct; constraint-based methods~\cite{ji2024testing,kumar2026localizing} fix the violated constraint where it surfaces rather than where it originates; and evolutionary methods~\cite{guo2026se} resample and recombine entire trajectories without locating the responsible step. What is missing is a trajectory-level mechanism that converts execution feedback into a localized update signal: one that identifies the earliest step at which execution contradicts the plan, and directs both reasoning and repair to that point.

To this end, we propose \textbf{PIVOT} (Plan--Inspect--eVOlve Trajectories), a self-supervised framework that formulates trajectory refinement as iterative optimization in a discrete, language-defined space. PIVOT defines a trajectory-level objective combining goal-achievement loss, plan--execution divergence, and execution cost. Since exact gradients are unavailable for language trajectories, PIVOT uses structured natural-language feedback as a surrogate \emph{textual gradient}~\cite{yuksekgonu24}, guiding local updates to the unsupported parts of the plan. A preference-based monotonic acceptance rule then keeps only refinements that improve the estimated trajectory loss, yielding a gradient-like optimization procedure over plans without requiring model fine-tuning. Inspired by backpropagation, PIVOT treats the executed trajectory as an unrolled computation graph and propagates the observed failure backward to the earliest decision whose planned assumption is contradicted by execution evidence. The subsequent update is localized accordingly: the validated prefix is retained, and only the downstream suffix is regenerated.

We evaluate PIVOT on two complementary benchmarks: DeepPlanning~\cite{zhang2026deepplanning}, which stresses long-horizon constraint satisfaction in Travel and Shopping Planning, and GAIA~\cite{mialon2023gaia}, which tests open-domain tool use and multi-step reasoning. PIVOT performs best on constraint-heavy planning tasks, where execution feedback exposes recoverable violations for trajectory-level repair. On GAIA, gains are more limited by retrieval quality, but PIVOT still improves task success in realistic tool-augmented settings. Across model families, PIVOT complements backbone scaling and remains token-efficient relative to competing refinement methods.

\section{Related Work}
\label{sec:related}

\textbf{Trajectory analysis and failure diagnosis.}
A growing body of work studies the reliability gap between planned and executed agent behavior through trajectory analysis and failure diagnosis~\cite{cemri2025multi, deshpande2025trail, ji2024testing, ma2026maestro, zhang2025agentracer, zhang2025agent, zhu2025llm}. AgentDebug~\cite{zhu2025llm} models agent behavior as a modular pipeline and shows how errors propagate across components, while VeriLA~\cite{sung2025verila} introduces human-centered criteria for identifying component-level failures. In multi-agent settings, MAST~\cite{cemri2025multi} and attribution benchmarks such as Who\&When~\cite{zhang2025agent} and AgenTracer~\cite{zhang2025agentracer} highlight the difficulty of localizing failures in long trajectories. Constraint-based methods~\cite{ji2024testing,kumar2026localizing} check plans against task requirements and correct the violating steps. These efforts show that failures can be localized, but either stop at diagnosis or check plans against static task requirements rather than against execution evidence. PIVOT closes this loop: the earliest step at which execution contradicts the plan becomes the starting point of repair.

\textbf{Iterative refinement and trajectory-level optimization.}
Recent self-critique methods~\cite{madaan2023selfrefine, shinn2023reflexion, Gou+24, bohnet2025enhancing} show that structured refinement of intermediate reasoning and tool use can yield strong empirical gains. However, such approaches remain operating through step-wise corrections rather than modeling the full trajectory as an optimizable object. PromptAgent~\cite{wang2023promptagent} similarly frames optimization as search over prompts and intermediate states, but does not directly optimize execution-grounded trajectory discrepancies. SE-Agent~\cite{guo2026se} moves closer to trajectory-level optimization by treating trajectories as genotypes whose phenotypic expression is problem-solving performance. This shares PIVOT's view of trajectories as optimizable objects, but derives its signal from \emph{cross-trajectory diversity and recombination} rather than from \emph{how intended plans diverge during execution}. PIVOT instead models the plan--execution discrepancy as a structured loss and interprets execution feedback as a textual gradient, enabling gradient-like optimization directly in trajectory space through a task-agnostic, self-supervised loop.

\begin{figure*}[htb]
    \centering
    \includegraphics[width=0.9\linewidth]{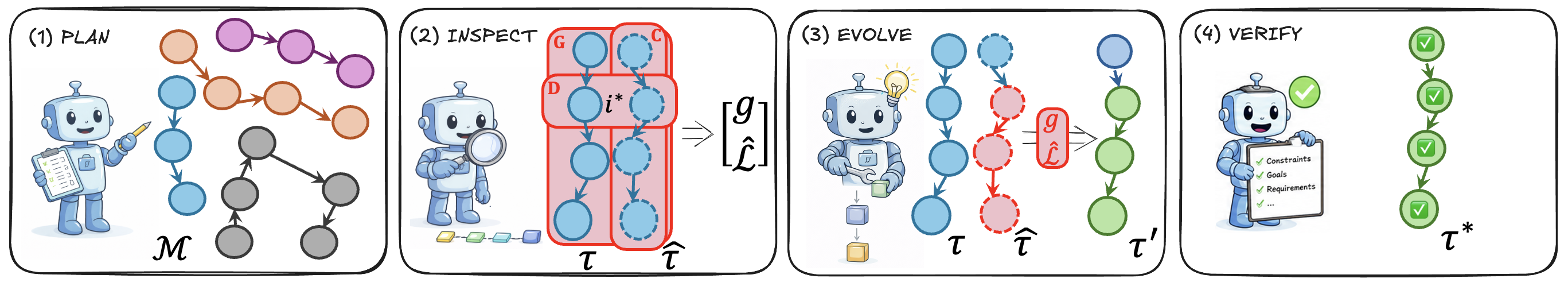}
    \caption{\textbf{Overview of PIVOT.} Given a task $t$, \textsc{Plan} states an explicit plan
    $\tau$, whose execution in $\mathcal{M}$ yields the trace $\hat{\tau}$.
    \textsc{Inspect} traces the unmet constraints $\widehat{\mathcal{L}}$
    backward through $\hat{\tau}$ to the earliest step $i^\star$ contradicted by
    execution evidence, producing a repair signal $g$. \textsc{Evolve} preserves
    the prefix before $i^\star$ and regenerates only the suffix, yielding
    $\tau'$. \textsc{Verify} checks the final trajectory $\tau^\star$ against all
    task constraints.}
	\label{fig:detailed_overview}
\end{figure*}

\section{PIVOT: Discrepancy-Driven Trajectory Refinement}
\label{sec:system}
\subsection{Problem Setup}
We consider a task $t$ to be completed over an environment $\mathcal{M} = (\mathcal{S}, \mathcal{A}, \mathcal{O})$, where $\mathcal{S}$ is the state space, $\mathcal{A}$ is the action space, $\mathcal{O}$ the space of outcomes after each step, and a transition function $f(\cdot)$ that maps state-action interactions to next states and outcomes , $f:\mathcal{M}\rightarrow \mathcal{S}\times \mathcal{O}$, reflecting the environment dynamics induced by agent actions.

A trajectory $\tau_K$ of length $K$ is defined as $\tau_K = \left((s_0, a_0), \dots, (s_K, a_K)\right)$, where $k\leq K$, $s_k \in \mathcal{S}$ and $a_k \in \mathcal{A}$. Executing $\tau_k$ up to a step $k$ produces an outcome $o_k$ evaluated as $G(\hat{\tau}_{\leq k}, o_k)$, with transitions following $(s_{k+1}, o_{k+1}) = f(s_k, a_k, o_k)$.

These definitions are task-agnostic: the initial plan $\tau^{(0)}$ should preserve user-specified requirements, the executed trace $\hat{\tau}^{(0)}$ reveals where they are no longer supported, \texttt{\textsc{Inspect}} produces a repair signal, and \texttt{\textsc{Evolve}} rewrites the remaining trajectory.

\subsection{PLAN: Trajectory Generation and Forward Rollout}
PIVOT begins with the \texttt{\textsc{Plan}} \;\includegraphics[height=1.4em]{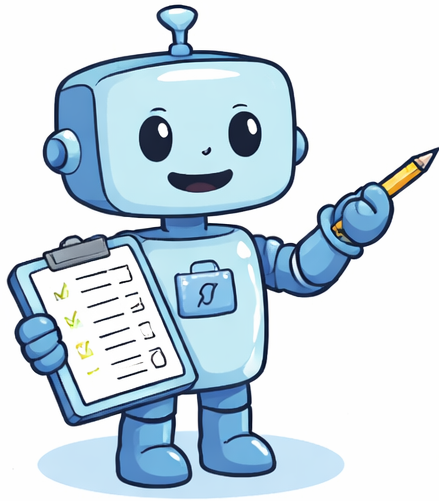} module, which converts the task into an explicit candidate trajectory. Like chain-of-thought reasoning, agent execution unfolds as an explicit reasoning-and-action trace. We view that trace as an unrolled forward computation in trajectory space. A planned trajectory $\tau$ induces an execution trace $\hat{\tau}$ when interacting with the environment, that can diverge due to incorrect assumptions or infeasible actions. We capture this mismatch through a trajectory-level objective:
\begin{equation}
\mathcal{L}(\tau) = \ell_G(\hat{\tau}, o_K) +  D(\tau, \hat{\tau}) +   C(\hat{\tau}),
\label{eq:TotalLoss}
\end{equation}
This objective combines three complementary sources of error: The term $\ell_G(\hat{\tau}, o_K)$ captures failure to achieve the desired task outcome, i.e.  a terminal goal-achievement loss derived from the reward-like success function $G(\hat{\tau}, o_K)$ as $\ell_G(\hat{\tau}, o_K) = 1 - G(\hat{\tau}, o_K)$,  $D(\tau, \hat{\tau})$ measures divergence between planned and executed trajectories, and $C(\hat{\tau})$ penalizes inefficient execution, defined as the number of tool calls:
$C(\hat{\tau}) = \sum_{i=0}^{K} \mathbf{1}[a_i \text{ invokes a tool}]$. 
This decomposition separates three different failure modes. The term $\ell_G(\hat{\tau}, o_K)$ rises when the final outcome fails to satisfy the task, $D(\tau, \hat{\tau})$ increases once execution no longer preserves the constraint structure encoded in the plan, and $C(\hat{\tau})$ captures the computation spent on trajectories that have already gone off course.

To localize failure, we identify the first step where planned and executed trajectories diverge:
\begin{equation}
i^\star = \min \{ i : D(\tau_{\leq i}, \hat{\tau}_{\leq i}) \geq T \},
\end{equation}
where $T$ is a divergence threshold. The index $i^\star$ marks the earliest step at which the executed trace no longer provides sufficient support for the planned trajectory prefix, and therefore identifies the point from which local repair should begin.

\subsection{INSPECT: Backward Localization over the Executed Trace}
The objective in Eq.~\ref{eq:TotalLoss} provides a unified signal for trajectory refinement by aggregating goal-level, trajectory-level, and efficiency-related errors. If Section~3.2 treats the agent trace as a forward rollout, then \texttt{\textsc{Inspect}}\;\includegraphics[height=1.4em]{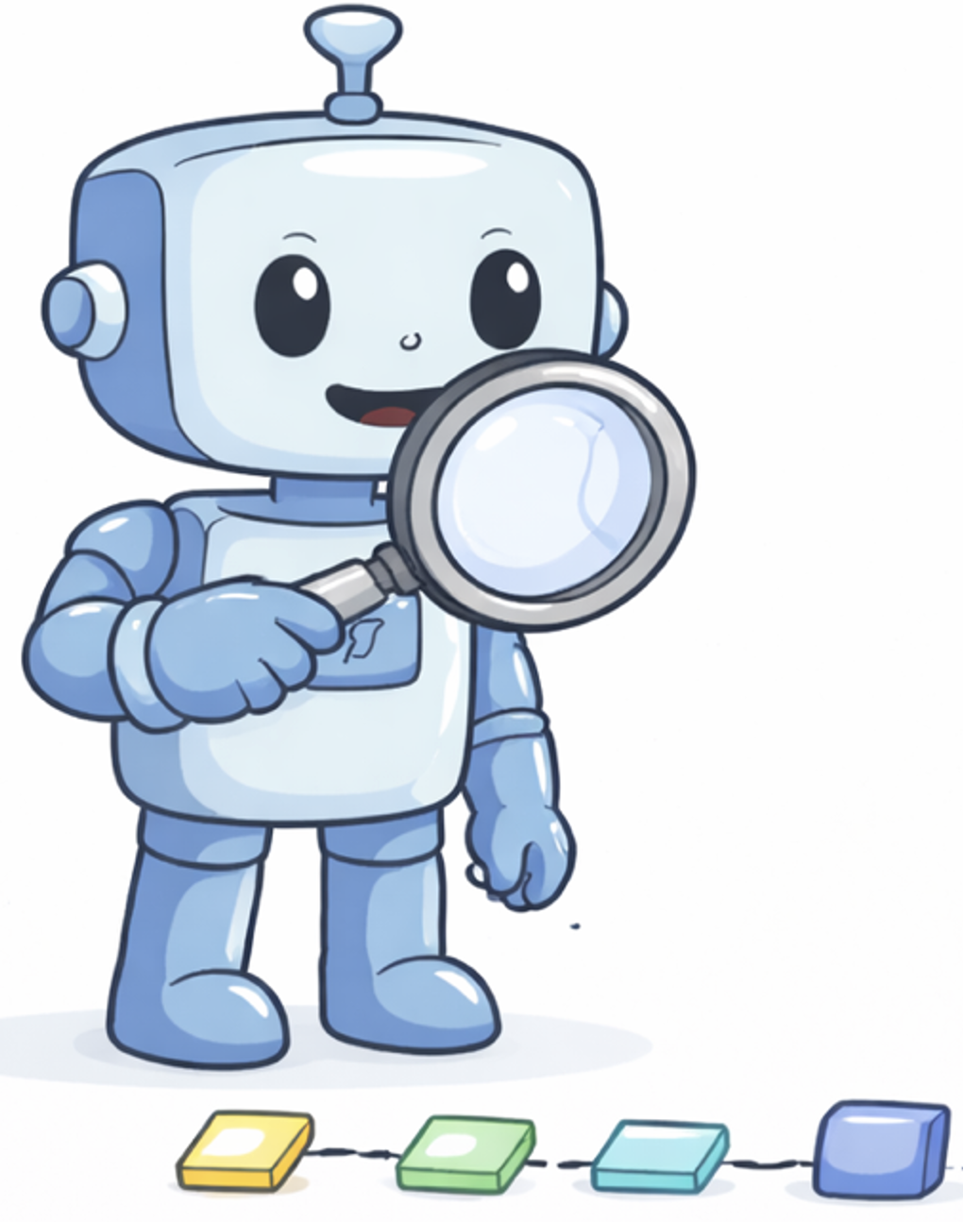} performs the reverse operation: once a task-level loss is observed, it analyzes the trajectory backward to identify the earliest causally responsible break. For a specific trajectory, the gradient of $\mathcal{L}$ becomes:
\begin{equation}
\frac{\partial \mathcal{L}}{\partial \tau}
\;\approx\;
\sum_{i \leq K}
\left[
\nabla_{\tau_i} \ell_G(\cdot)
+
\nabla_{\tau_i} D(\cdot)
+
\frac{\partial C(\cdot)}{\partial \tau_i}
\right],
\label{eq:back_propagation}
\end{equation}

Each term contributes error signals to every trajectory step, playing backward credit assignment over the discrete trace, traversing dependencies backward from each violation and stopping at the
earliest step contradicted by execution evidence. It returns $(\hat{\imath}^\star,g)$, where the repair signal $g$ states the observed failure, its manifestation in the trace, the break at $\hat{\imath}^\star$, and what to do differently from there. 

\subsection{EVOLVE: Local Trajectory Repair}
Given a backward attribution signal from \texttt{\textsc{Inspect}}, the role of \texttt{\textsc{Evolve}} \;\includegraphics[height=1.4em]{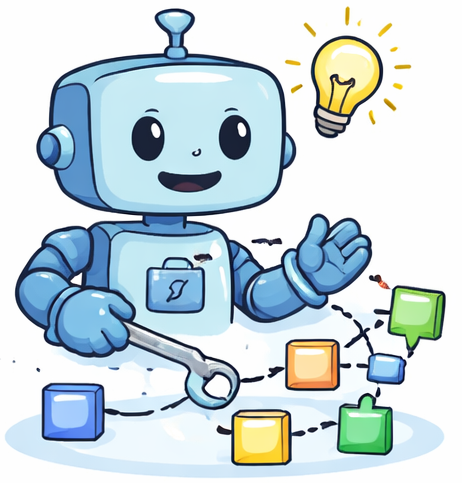} is to produce an improved trajectory by modifying  current plan in response to identified discrepancies. At iteration $r \leq R$, \texttt{\textsc{Evolve}} proposes an updated trajectory:
\begin{equation}
    \tau' = \texttt{\textsc{Evolve}}(\hat{\tau}^{(r)}, \tau^{(r)}, g^{(r)}),
\end{equation}
where $g^{(r)}$ specifies both the location and nature of the failure. Rather than regenerating the full trajectory, the update operates locally: the validated prefix $\tau_{\leq i^\star}$ is preserved, while the unsupported suffix is rewritten, focusing  on the part of the trajectory responsible for the current loss.

The candidate trajectory is then executed and evaluated:
\begin{equation}
    (\widehat{\mathcal{L}}', g') = \texttt{\textsc{Inspect}}(\hat{\tau}', \tau', t).    
\end{equation}

The update is accepted only if it improves the objective:
\begin{equation}
    \tau^{(r+1)} =
    \begin{cases}
    \tau' & \text{if } \widehat{\mathcal{L}}' \prec \widehat{\mathcal{L}}^{(r)}, \\
    \tau^{(r)} & \text{otherwise}.
    \end{cases}    
\end{equation}

This accept/reject mechanism ensures monotonic improvement, resembling a gradient-based update in the trajectory space, where \texttt{\textsc{Inspect}} provides a structured approximation of the loss and \texttt{\textsc{Evolve}} performs a local step guided by this signal.

\subsection{VERIFY: Global Constraint Validation.}
While \texttt{\textsc{Inspect}} provides local feedback for trajectory refinement, it does not guarantee that the final trajectory fully satisfies the original task specification. To address this, PIVOT applies a final validation step using \texttt{\textsc{Verify}} \;\includegraphics[height=1.4em]{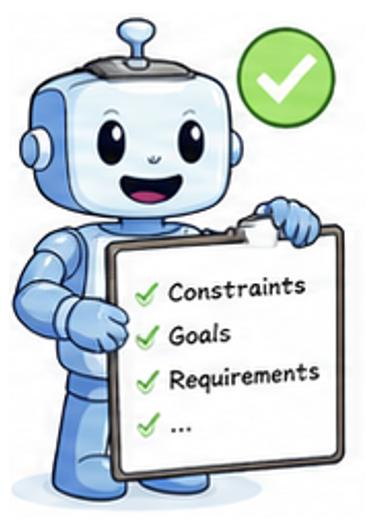}, which performs a global check over the complete trajectory.
Given a candidate solution $\tau$, \texttt{\textsc{Verify}} evaluates whether all user-specified constraints and requirements are satisfied. This step operates independently of the trajectory-level discrepancy signals used during refinement, ensuring that improvements in intermediate behavior do not come at the expense of violating the original task.
The final output of the system is therefore obtained by applying:
\begin{equation}
    \tau^\star = \texttt{\textsc{Verify}} \left( \operatorname*{arg\,min}_{r \leq R}^{\prec} \ \widehat{\mathcal{L}}^{(r)}, \ t \right),
\end{equation}
which enforces that the selected trajectory is both optimal under the refinement objective and consistent with the task specification.

\section{Experiments}
\label{sec:experiment}
We evaluate PIVOT on three axes: (1) \textbf{trajectory alignment optimization}: can explicitly optimizing the alignment between planned trajectories and executed behavior translate into state-of-the-art task performance and improved reliability on long-horizon agent benchmarks? (2) \textbf{execution efficiency}: are these gains practically attainable under favorable token usage and execution overhead? and (3) \textbf{improvement attribution}: which components drive the observed gains, and what gives PIVOT an advantage over prior agent frameworks?

\subsection{Experimental Setup}

\textbf{Benchmarks and evaluation Metrics.}
We evaluate PIVOT on two complementary benchmarks. DeepPlanning~\cite{zhang2026deepplanning} is a long-horizon planning benchmark with two tool-grounded domains: \emph{Travel Planning} and \emph{Shopping Planning}, each containing $120$ tasks. We follow the official benchmark protocol for the main evaluation metrics. 
GAIA~\cite{mialon2023gaia} contains $466$ human-authored, open-domain questions that require multi-step reasoning and tool use in realistic assistant settings. We report exact-match task success rate as guided in the original dataset paper. We equip the agent with four basic tools in the evaluation: \texttt{web\_search}, \texttt{web\_fetch}, \texttt{read\_file}, and \texttt{python\_exec}. More details for the dataset and tool setups are in App.~\ref{app:gaia_tools}.

\textbf{Experimental settings.}
We instantiate all methods on the same set of underlying model backbones, including Qwen3-235B~\cite{qwen3technicalreport}, Claude 4.5 model family, and Claude 4.6 Opus. All methods share the same benchmark tools, system prompts, and task environments, and baselines are re-implemented within this common harness, as listed in the App.~\ref{appendix:prompts}.

\begin{table*}[htb]
\centering
\small
\setlength{\tabcolsep}{4.0pt}
\renewcommand{\arraystretch}{1.12}
\resizebox{\textwidth}{!}{%
\begin{tabular}{@{}lccccc@{\hspace{8pt}}ccccc@{\hspace{8pt}}ccccc@{}}
\toprule
& \multicolumn{5}{c}{\textbf{Travel Planning}} & \multicolumn{5}{c}{\textbf{Shopping Planning}} & \multicolumn{5}{c}{\textbf{GAIA}} \\
\cmidrule(lr){2-6}\cmidrule(lr){7-11}\cmidrule(l){12-16}
\textbf{Method} &
\twoline{\textbf{Q3}}{\textbf{235B}} &
\twoline{\textbf{Haiku}}{\textbf{4.5}} &
\twoline{\textbf{Sonnet}}{\textbf{4.5}} &
\twoline{\textbf{Opus}}{\textbf{4.5}} &
\twoline{\textbf{Opus}}{\textbf{4.6}} &
\twoline{\textbf{Q3}}{\textbf{235B}} &
\twoline{\textbf{Haiku}}{\textbf{4.5}} &
\twoline{\textbf{Sonnet}}{\textbf{4.5}} &
\twoline{\textbf{Opus}}{\textbf{4.5}} &
\twoline{\textbf{Opus}}{\textbf{4.6}} &
\twoline{\textbf{Q3}}{\textbf{235B}} &
\twoline{\textbf{Haiku}}{\textbf{4.5}} &
\twoline{\textbf{Sonnet}}{\textbf{4.5}} &
\twoline{\textbf{Opus}}{\textbf{4.5}} &
\twoline{\textbf{Opus}}{\textbf{4.6}} \\
\midrule
ReACT~\cite{yao2023react} & 28.3 & 35.2 & 53.1 & 62.9 & 86.0 & 10.8 & 25.8 & 40.8 & 43.3 & 50.8 & 31.5 & 37.0 & 40.6 & 59.4 & 57.0 \\
Self-Critique~\cite{bohnet2025enhancing} & 30.1 & 45.1 & 52.6 & 75.2 & 87.3 & 16.7 & 35.8 & \resultsecond{55.8} & \resultsecond{53.3} & \resultsecond{70.0} & \resultsecond{37.6} & \resultbest{44.8} & \resultsecond{51.5} & \resultsecond{64.2} & \resultsecond{68.5} \\
SE-Agent~\cite{guo2026se} & 29.6 & 41.9 & 52.6 & 63.9 & 82.8 & 15.8 & 31.7 & 45.8 & 43.7 & 53.7 & 37.5 & 37.0 & 40.0 & 57.6 & 56.4 \\
AgentDebug~\cite{zhu2025llm} & 25.1 & 42.7 & 46.2 & 59.3 & 86.1 & 15.0 & 33.3 & 47.5 & 45.8 & 53.3 & 33.4 & \resultsecond{41.8} & 47.9 & 63.6 & 66.7 \\
\midrule
\rowcolor{gray!8}
\pivotmethod{PIVOT (\emph{self-judged})} & \resultsecond{39.8} & \resultsecond{53.7} & \resultsecond{59.8} & \resultsecond{76.5} & \resultsecond{89.3} & \resultsecond{20.0} & \resultsecond{37.5} & 43.3 & 48.3 & 55.8 & 35.2 & 40.6 & 41.8 & 60.6 & 64.8 \\
\rowcolor{gray!14}
\pivotmethod{PIVOT (\emph{oracle})} & \resultbest{39.9} & \resultbest{57.3} & \resultbest{67.1} & \resultbest{82.7} & \resultbest{96.2} & \resultbest{76.7} & \resultbest{89.2} & \resultbest{89.2} & \resultbest{94.2} & \resultbest{98.3} & \resultbest{41.8} & \resultbest{44.8} & \resultbest{52.7} & \resultbest{68.5} & \resultbest{71.5} \\
\bottomrule
\end{tabular}%
}

\caption{\textbf{Main results across DeepPlanning and GAIA.} Travel reports composite score, Shopping reports case accuracy, and GAIA reports task success rate (all in \%). Each column group corresponds to one benchmark, and each row reports the same agent framework instantiated on the same set of backbones. We bold the best result and underline the second-best result within each column.}
\label{tbl:main_results}
\end{table*}

\begin{figure*}[htb]
    \centering
    \includegraphics[width=0.9\linewidth]{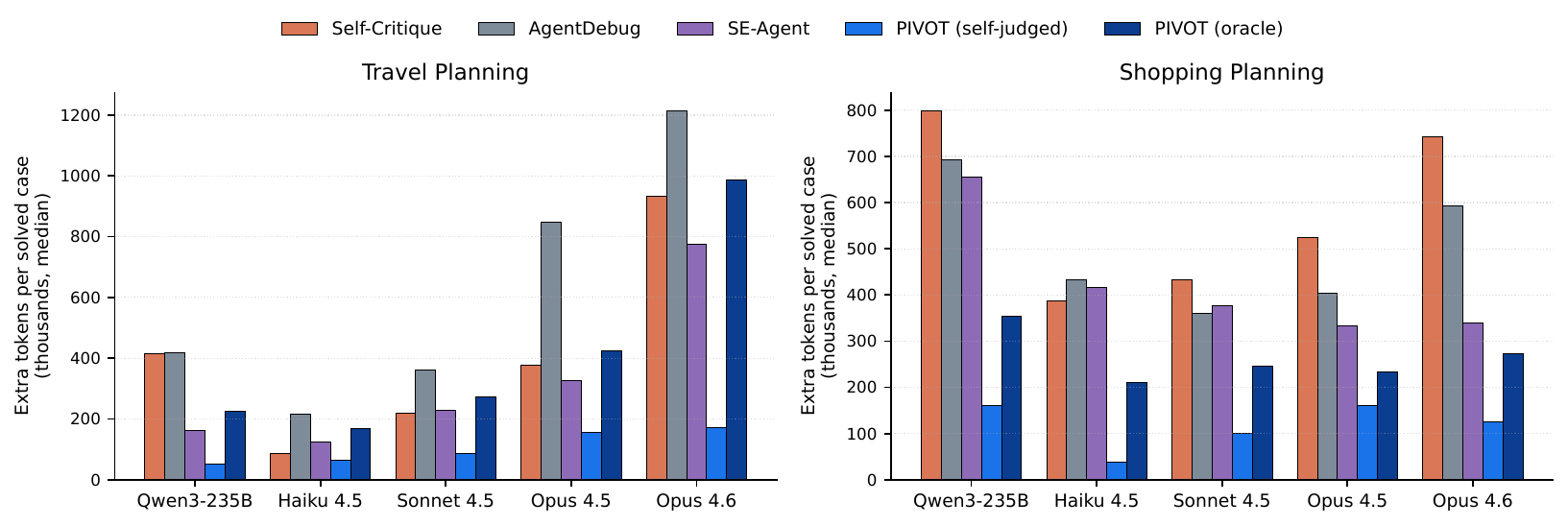}
    \caption{\textbf{Extra token cost per solved case, relative to  ReACT baseline.}
    Bars report  median extra billed tokens (input + output) on solved cases (composite score $\geq 0.7$); lower is better. PIVOT (w/o HITL) adds 3--5$\times$ fewer tokens than Self-Critique, SE-Agent or AgentDebug; the HITL variant remains cheaper than either baseline. Since autoregressive decoding time grows with the number of generated tokens, token usage serves as a proxy for both inference cost and latency.}
    \label{fig:extra_tokens}
\end{figure*}

\textbf{Feedback regimes.} We evaluate PIVOT under two feedback regimes differ in the source of the unmet-constraint set $\widehat{\mathcal{L}}$. In the default \emph{self-judged} regime, the agent determines $\widehat{\mathcal{L}}$ itself through \textsc{Verify} without external signal. In the \emph{oracle} regime, $\widehat{\mathcal{L}}$ is supplied by the benchmark's deterministic and rule-based evaluator. The reports only contains which constraints are unmet (e.g., ``hotel price outside the requested range'') but never how to satisfy them to mimic the real human interaction with agent. No human annotators are involved. Because this signal is exact and constraint-level, the oracle regime is an upper bound rather than a deployable setting, and it receives strictly more information than any baseline.

\textbf{Baselines.}
We compare PIVOT against four representative agent baselines. ReACT~\cite{yao2023react} serves as a canonical single-pass reasoning-and-acting baseline. Self-Critique~\cite{bohnet2025enhancing} and SE-Agent~\cite{guo2026se} are self-refinement baselines, providing competitive benchmarks for iterative plan improvement. AgentDebug~\cite{zhu2025llm} is a debugging framework that isolates root-cause failures and generates corrective feedback for recovery, as a relevant comparison for our discrepancy-driven refinement setting.

ReAct~\cite{yao2023react} and SE-Agent~\cite{guo2026se} use no external signal, matching self-judged PIVOT. Self-Critique~\cite{bohnet2025enhancing} and AgentDebug~\cite{zhu2025llm} baselines rely on human-provided or externally supplied feedback in their evaluated setups. We therefore compare them with oracle PIVOT. Since only AgentDebug releases usable code, we re-implement all baselines under identical tools, prompts, and state-of-the-art backbones rather than the weaker models of their original papers, a setting that favors the baselines.

\begin{figure*}[htb]
    \centering
    \includegraphics[width=0.9\linewidth]{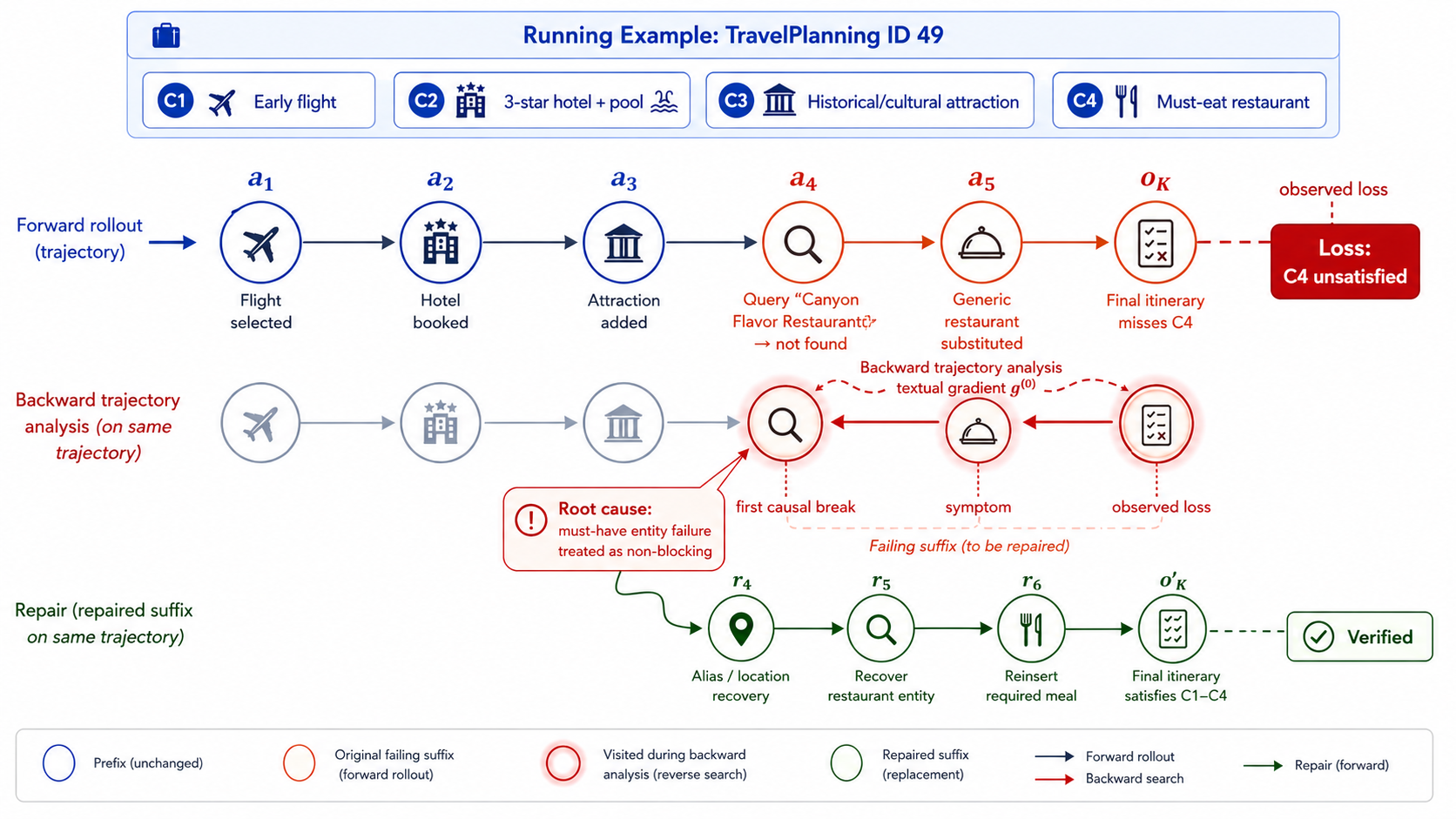}
    \caption{\textbf{Trajectory-level repair via PIVOT with HITL feedback.} 
The initial rollout violates constraint \(C_4\) after a failed retrieval for ``Canyon Flavor Restaurant,'' producing an invalid itinerary. \texttt{\textsc{Inspect}} localizes the first causal break and produce a textual gradient; \texttt{\textsc{Evolve}} repairs only the unsupported suffix while preserving the valid prefix. The refined trajectory restores the required entity and satisfies \(C_1\)--\(C_4\). }
	\label{fig:pivot_sample}
\end{figure*}

\subsection{Main Results}

Table~\ref{tbl:main_results} summarizes the main evaluation results. We have three key observations:

\textbf{PIVOT is consistently strong across models and benchmarks.} Because the baselines receive different inputs, we compare PIVOT with each group at its level of information. At
equal information, self-judged PIVOT improves over ReAct in all 15 settings and over the stronger of ReAct and SE-Agent in 13. The gains are largest on Travel Planning, where PIVOT is best among all methods on every backbone, and they persist on the strongest backbone so trajectory repair complements backbone scaling. Against baselines that know which trajectories failed with precise feedback, self-judged PIVOT still wins 7 of 15, and with constraint-level feedback, PIVOT is best or tied in all. Under equal-input comparison, PIVOT leads uniformly, indicating its robust performance.

\textbf{Retrieval bounds the gains on GAIA.} All methods on GAIA use the same
minimal four-tool setup, so absolute numbers are not comparable to papers with richer tooling. GAIA failures typically stem from evidence that was never retrieved rather than evidence that was
misused; the trajectory then contains no unsupported step to repair from, and re-querying the same tools returns the same results. PIVOT still improves over baselines on all five backbones, but by smaller margins than on DeepPlanning, which delineates where execution-grounded repair is most effective. 

\textbf{PIVOT is token-usage efficient.} To isolate the computational cost, we compute the extra tokens each method spends on cases it solves relative to the ReACT baseline in DeepPlanning. We report the per-case median and aggregate across backbones with the median-of-medians to be robust to long-tail outliers where weaker backbones enter tool-call loops under ReACT. 
In Fig.~\ref{fig:extra_tokens}, PIVOT saves 345k (4.2×) and 320k (4.1×) extra tokens per solved case compared to Self-Critique and AgentDebug due to: (1) Both baselines require a second full trajectory conditioned on a long failure transcript, scaling their cost with how much the backbone needs to explore in the first place. PIVOT instead steers a single trajectory with four in-line prompts, adding a bounded overhead that is largely independent of backbone verbosity. (2) PIVOT bounds the worst case: when a backbone enters a long, non-productive exchange, PIVOT inspect/evolve triggers can redirect it within the current trajectory, whereas post-hoc methods inherit the full failed transcript and must re-run for it.

\vspace{-2mm}
\begin{table*}[t]
\centering
\newcommand{\drop}[1]{%
  \textcolor{red!80!black}{$\downarrow$\,#1}%
}
\resizebox{\linewidth}{!}{%
\begin{tabular}{@{}l cccc ccccc r@{}}
\toprule
& \raisebox{-0.6em}{\includegraphics[height=2.2em]{icons/plan_robot.png}}
& \raisebox{-0.6em}{\includegraphics[height=2.2em]{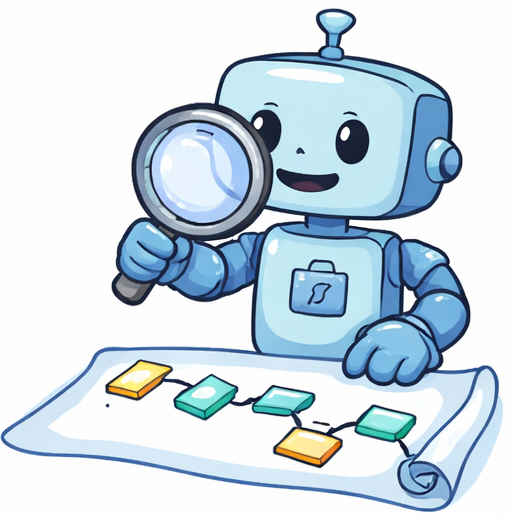}}
& \raisebox{-0.6em}{\includegraphics[height=2.2em]{icons/evolve_robot.png}}
& \raisebox{-0.6em}{\includegraphics[height=2.2em]{icons/verify_robot.png}}
& \textbf{Q3-235B}
& \textbf{Haiku 4.5}
& \textbf{Sonnet 4.5}
& \textbf{Opus 4.5}
& \textbf{Opus 4.6}
& \textbf{Avg. Drop} \\
\midrule
ReAct~\cite{yao2023react}
& $-$ & $-$ & $-$ & $-$
& 28.33 & 35.23 & 53.07 & 62.86 & 86.02
& \textcolor{gray}{---} \\
\midrule
\multirow{5}{*}{\rotatebox[origin=c]{90}{\textsc{PIVOT}}}
& \Checkmark & \Checkmark & \Checkmark & \Checkmark
& 39.81 & 53.69 & 59.82 & 76.45 & 89.27
& \textcolor{gray}{---} \\
& \xmark   & \Checkmark & \Checkmark & \Checkmark
& 15.34 & 51.51 & 54.69 & 62.79 & 77.89
& \drop{11.36} \\
& \Checkmark & \xmark   & \Checkmark & \Checkmark
& 29.40 & 50.18 & 58.93 & 71.51 & 87.92
& \drop{4.22} \\
& \Checkmark & \Checkmark & \xmark   & \Checkmark
& 20.57 & 50.62 & 54.74 & 61.28 & 77.92
& \drop{10.78} \\
& \Checkmark & \Checkmark & \Checkmark & \xmark
& 18.78 & 44.74 & 52.03 & 60.62 & 76.43
& \cellcolor{red!12}\drop{13.29} \\
\bottomrule
\end{tabular}%
}
\caption{\textbf{Ablation of PIVOT stages on Travel Benchmark.} 
Each row removes one component (\xmark) while keeping others (\Checkmark); Avg. Drop reports the mean degradation relative to the full pipeline. 
All stages contribute to performance, confirming that PIVOT operates as a tightly coupled refinement loop rather than a set of independent modules. 
\textsc{Verify} has the largest impact, highlighting the importance of final constraint validation, followed by \textsc{Plan} and \textsc{Evolve}, which govern trajectory initialization and repair. \textsc{Inspect} shows a smaller but consistent contribution, indicating that even imperfect discrepancy signals remain useful for guiding updates.}
\label{tab:ablation_study}
\end{table*}

\subsection{A Trajectory-Level View of Failures and Repair}
\label{sec:trajectory_view}
We visualized a running example to illustrate how PIVOT locates break point and repairs as shown in Figure~\ref{fig:pivot_sample}. In this section, we examine in general about where errors arise along
the trajectory, and how the scope of repair affects its outcome based on the trajectories from the Travel Planning.

\textbf{Errors surface late but become irrecoverable early.} The step at which an error becomes visible differs from the step at which it becomes irrecoverable. We probe these two points separately on failed trajectories. For visibility, we leverage a rule-based check 
to flag the conflicts against each user request and tool outputs. For irrecoverability, we use a resumption probe: we keep each step $k$ verbatim including its tool outputs, and let a fixed model continue the task without feedback. The resumer can satisfy a violated constraint only if the retained prefix has not yet committed to the error, so the last $k$ at which resumption succeeds marks where the error is locked in.

The two probes locate errors at different points. The rule-based check finds a conflict in 27 of 112 failed trajectories. In all 27 the wrong value is first written at the final assembly step, although the correct value had been retrieved earlier. The resumption probe labels 20 cases, and places the lock-in point mid-trajectory in 14, at the first step in 4, and at the last step in only 2. Together they reproduce the pattern of Figure~\ref{fig:teaser}: the violation is written during plan assembly, but the prefix that makes it unavoidable is fixed several steps earlier. Locating this point precisely is inherently hard: a single-sample probe with Claude 4.5 Opus as the resumer agrees with the majority label within one step in 63\% of cases. Existing diagnostics are not designed for this difficulty and fail in opposite directions. Across 874 cases, 57.3\% of AgentDebug diagnoses fall into a single category and 60.4\% blame the first step, which is too coarse to locate the lock-in point. SE-Agent's critic over-flags instead: its precision drops from 0.99 to 0.49 as the backbone strengthens, and 40 of its 95 false-positive corrections on Claude 4.6 Opus strictly lower the score. PIVOT therefore does not rely on pinpointing this step exactly; instead, it repairs conditioned on the executed prefix.

\textbf{Repair must be local and actionable.} Effective repair must state what to change and keep what execution has already validated; existing
methods violate one or the other. SE-Agent's critique triggers broad
re-plans that fix one element while breaking others: a suboptimal flight
prompts a rewrite that loses attraction coverage, meal timing, or transfer
feasibility. AgentDebug's taxonomy instead yields a verdict rather than a
repair, as 53\% of its correction strings merely restate the predicted label. PIVOT's diagnosis names the violated constraint, its manifestation in the trace, the step where support was lost, and what to do differently, and is applied on top of the executed trajectory. With the diagnosis fixed, this outperforms replanning from the query by 5.6 points. Repair succeeds when the evidence was retrieved but ignored, and also when the root cause lies far upstream, e.g., a train fabricated at the first query. It fails when the evidence cannot be retrieved at all, which bounds the gains on GAIA. Local repair can also leave a plan whose steps are each supported but whose total violates a global constraint such as budget, which is why \textsc{Verify} checks the assembled plan as a whole.

\subsection{Why PIVOT Works}
\label{sec:ablation}

\begin{figure}[h]
    \centering
    \includegraphics[width=\linewidth]{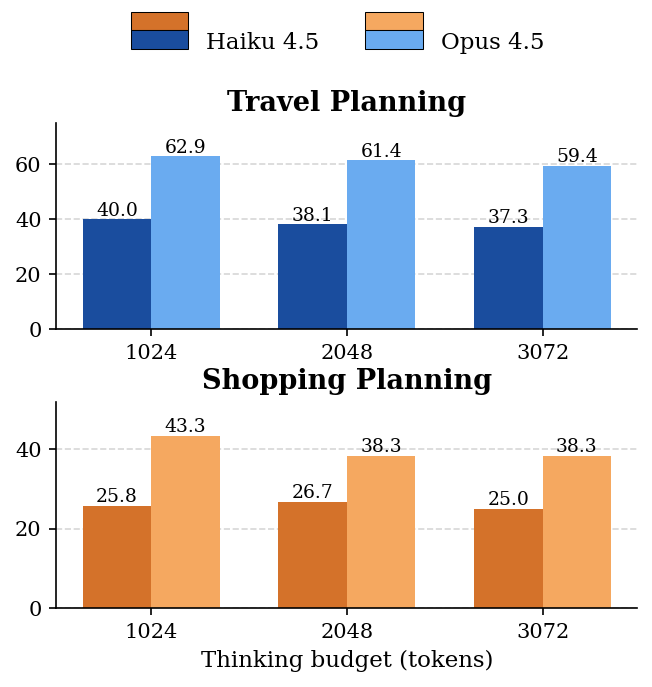}
    \caption{\textbf{Thinking-budget ablation.} Scaling the extended-thinking budget from 1024 to 3072 tokens does not improve performance on either backbone or benchmark.}
    \label{fig:thinking_budget}
\end{figure}

\textbf{Thinking is in the right place.} 
A natural alternative to structured reflection is to simply allocate more test-time compute by raising the extended-thinking budget but this does not work, as in Figure~\ref{fig:thinking_budget}; increasing the budget yields no consistent gain on either benchmark. Inspection of the trajectories reveals that under the default extended-thinking regime, 100\% of thinking blocks fire on the first assistant turn, spending tokens on task decomposition and tool selection, while 99.2\% of final-plan generation steps produce \textit{zero} thinking tokens. The model actually uses only $\sim$230 thinking tokens on average, well below the 1024 ceiling, and raising the ceiling does not redirect where that thinking happens. The hardest part of the task, synthesizing 20+ tool outputs into a coherent multi-day itinerary under interacting constraints, is executed with no reasoning at all, regardless of budget. PIVOT addresses this distributional problem rather than the quantity problem: its plan--inspect--evolve--verify schedule triggers reasoning at multiple points along the trajectory. After tool returns, before plan assembly, and before the final answer, thinking is allocated precisely where synthesis and verification happen, not only at the upfront planning step.

\textbf{Ablation study on importance of PIVOT components.} Table~\ref{tab:ablation_study} reports the effect of disabling each PIVOT stage while keeping the remaining stages active. The largest degradation occurs when removing \texttt{\textsc{Verify}} ($-13.3$ avg.), followed by \texttt{\textsc{Plan}} ($-11.4$ avg.), \texttt{\textsc{Evolve}} ($-10.8$ avg.), and \texttt{\textsc{Inspect}} ($-4.2$ avg.). This suggests that PIVOT's gains do not come solely from better upfront planning; instead, reasoning at the end of the trajectory, where the draft plan is checked against the original constraints, contributes the most. The effect persists even for stronger backbones: disabling \texttt{\textsc{Verify}} reduces Opus~4.6 by $12.9$ points, indicating that even highly capable models can silently emit constraint violations unless explicitly prompted to validate them. This finding is consistent with the thinking-allocation analysis above. Vanilla extended thinking spends most of its reasoning budget on the opening turn and produces little or no thinking during plan synthesis or final answering. \texttt{\textsc{Verify}} places reasoning precisely at this under-covered stage, where constraint violations are most likely to surface in the final output. Its benefit therefore comes from improved reasoning coverage rather than simply adding more reasoning tokens. \texttt{\textsc{Plan}} and \texttt{\textsc{Evolve}} also remain important, especially for weaker backbones, showing that both strong trajectory initialization and discrepancy-guided repair are necessary for robust performance.

\textbf{Why \textsc{Verify} matters most.} \textsc{Verify} is the only stage that reasons at plan assembly, and plan assembly is where errors are committed. \textsc{Plan}, \textsc{Inspect}, and \textsc{Evolve} act before or between tool calls, while the agent is still gathering evidence; the violations themselves are written later, when the agent combines that evidence into the final itinerary. In all 27 failed trajectories where a plan value contradicts a tool output, the correct value had been retrieved and the wrong one first appears at this assembly step, where the default agent performs no reasoning as we reported in Sec.~\ref{sec:trajectory_view}.

%

\section{Conclusions}
\label{sec:conclusions}
We studied where long-horizon LLM agents fail and how to repair them.
Errors are written at plan assembly, where models reason least, but become
irrecoverable several steps earlier, so correcting the final output does not reach their cause. PIVOT uses an explicit plan as the reference for checking execution, places reasoning where execution evidence must be reconciled with the plan, and repairs failures from the localized step on top of the executed trajectory. Empirical results across DeepPlanning and GAIA show that PIVOT is especially effective in long-horizon, constraint-driven settings where failures stem from recoverable plan--execution discrepancies. Overall, our findings position discrepancy-driven trajectory optimization as a principled mechanism for improving reliability in autonomous agents, particularly when constraint satisfaction, execution feedback, and iterative repair are central to task success.

\section{Limitations}
PIVOT proposes safer autonomous systems and interpretable failure attribution. However, these same capabilities introduce risks around reduced human oversight, dual-use potential, misplaced trust, and labor displacement that must be proactively addressed. The responsible path forward requires coupling PIVOT technical contributions with deployment guardrails, and ongoing evaluation in real-world settings beyond controlled benchmarks.

\newpage


\bibliography{custom}

\appendix

\newpage
\section{Prompts and Flow Discussion}
\label{appendix:prompts}

\subsection{Stage 1: PLAN - Structured Planning Before Action}

{\bf Intuitive Analogy}: Imagine a chef who just received an order for a multi-course dinner. Before touching any ingredients, they sit down with pen and paper: What dishes are we making? What ingredients do we need? Which stations handle what? What's the cooking sequence? What if the oven breaks or we're out of a key ingredient? This is what PLAN forces the AI agent to do before it touches any tool.

{\bf Core Mechanism}: 
\begin{itemize}
    \item Forces the agent to produce an explicit, structured plan before any tool is called
    \item The plan must include: goal restatement, information needs, tool selection per step, step dependencies, and failure contingencies
    \item Serves as the "blueprint" that all subsequent stages reference
\end{itemize}
\begin{promptbox}
    PLAN BEFORE ACT:\\\\
    Before using any tools, you MUST first output a plan inside <plan>...</plan> tags:
    \begin{enumerate}
        \item  Restate the user's goal in one sentence. 
        \item List each piece of information you need to gather, as numbered steps.
        \item For each step, note which tool you intend to use and what a successful result looks like.
        \item Identify dependencies between steps (e.g., "step 3 requires the URL from step 2").
        \item Note any potential failure modes (e.g., "if the search returns no results, try alternative query X").
    \end{enumerate}
After the plan, begin executing it step by step. Refer back to your plan as you go.
\end{promptbox}

{\bf Trigger Condition}: Appended to the system prompt. Runs once, at the very start of the task.

{\bf Design Rationale}: Research consistently shows that LLMs perform better when they "think before acting." Without an explicit planning step, agents tend to call tools reactively, grabbing whatever seems relevant in the moment rather than following a coherent strategy. By requiring the agent to articulate its plan upfront (including failure contingencies), we front-load the strategic thinking that prevents downstream waste.

\subsection{Stage 2: INSPECT — Periodic Reflection Checkpoints}

{\bf Intuitive Analogy}: Now the chef is actively cooking. After finishing each course, they pause to taste and assess: "Did this sauce reach the right consistency? Am I still on schedule for the next course? Does this plating match my original vision, or do I need to adjust my approach for the remaining dishes?" They don't wait until all courses are plated—they catch issues while there's still time to correct course.

{\bf Core Mechanism}:
\begin{itemize}
    \item Executes plans step-by-step in the actual environment
    \item Tracks discrepancies between what was planned and what actually happened
    \item  Identifies constraint violations, infeasible actions, and early warning signs
    \item  Generates structured diagnostic reports documenting exactly where and why things went wrong
    \item  Proactively detects problems rather than waiting for catastrophic failures
\end{itemize}

\begin{promptbox}
    REFLECT AFTER TOOL: \\\\
    Before your next action, briefly answer these three questions in 1-2 sentences each:
    \begin{enumerate}
        \item RESULT CHECK: Did the tool return what you expected? If not, what went wrong?
        \item PLAN STATUS: Which steps of your plan are complete, and which remain?
        \item REVISION: Do you need to change your approach for the remaining steps? If a search failed, what alternative query or source could you try?
    \end{enumerate}
    Then continue executing your plan.
\end{promptbox}

{\bf Trigger Condition}: Injected automatically every 3 tool-call rounds during execution.

{\bf Design Rationale}: Without periodic reflection, agents exhibit "tunnel vision", they keep executing the original plan even when early results clearly indicate it won't work. The 3-round interval is a deliberate balance: frequent enough to catch problems early, but not so frequent that it slows down execution with excessive self-reflection. The three-question structure (Result → Status → Revision) mirrors the observe-orient-decide loop used in real-world decision-making frameworks.

\subsection{Stage 3: EVOLVE — Adaptive Plan Revision on Failure}

{\bf Intuitive Analogy}: The chef discovers mid-service that a key ingredient is spoiled. Rather than just trying to salvage that same ingredient or make the dish anyway, they step back to diagnose: Why won't this work? Is there a completely different ingredient that provides the same flavor profile? Could I pivot to an entirely different dish that uses what's available? And if no substitution works, what's the best menu I can serve with the ingredients I do have? EVOLVE forces the agent through this same discipline: diagnose root cause, pivot to a fundamentally different strategy, or gracefully degrade—but never blindly retry the same failed approach.

{\bf Core Mechanism}:
\begin{itemize}
    \item Triggered specifically when a tool call fails or returns useless results
    \item Forces the agent to diagnose the root cause, propose a completely different approach (not just a retry), and identify a fallback if no alternative exists
    \item Capped at 3 invocations per task to prevent infinite retry loops
\end{itemize}

\begin{promptbox}
PLAN REVISION:\\

Your last tool call did not return useful results. Before retrying, you MUST:
\begin{enumerate}
    \item DIAGNOSE: Why did it fail? (wrong query terms? page not accessible? data not in this source?)
    \item ALTERNATIVE: Name a completely different approach to get this information. Do not repeat the same query.
    \item FALLBACK: If no alternative source exists, what partial answer can you construct from what you already have?
\end{enumerate}

Then execute your revised approach.
\end{promptbox}

{\bf Trigger Condition}: injected when a tool call fails, max 3 times

{\bf Design Rationale}: The most common failure pattern in AI agents is the "retry loop": when a search fails, the agent tries the same query (or a trivially different one) repeatedly. EVOLVE breaks this pattern by explicitly prohibiting repeated queries and requiring the agent to name a completely different approach. The 3-invocation cap is a cost-control mechanism: if three fundamentally different approaches all fail, the information likely isn't accessible, and the agent should construct the best partial answer it can.

\subsection{Stage 4: VERIFY — Final Quality Gate}

{\bf Intuitive Analogy}: All courses are plated and ready. Before sending them to the dining room, the chef does a final inspection against the original order: \checkmark Are all requested courses present? \checkmark Are dietary restrictions honored? \checkmark Is the presentation what was promised? \checkmark Is the temperature correct for each dish? If anything's missing or wrong, they fix it now. If something's impossible with available ingredients, they communicate it clearly to the customer. VERIFY is the agent's version of this final quality check before serving.

{\bf Core Mechanism}:
\begin{itemize}
    \item Runs once, just before the agent produces its final answer
    \item Forces the agent to re-read the original question and check its answer against every stated requirement
    \item Each requirement is explicitly marked as \checkmark\ satisfied or \xmark\  not satisfied
    \item Any gaps are either fixed on the spot or transparently acknowledged
    \item Addresses root causes identified through backward analysis, and then fundamentally reconstructs plans rather than just patching individual steps
    \item Produces plans that are more executable and constraint-aware
\end{itemize}

\begin{promptbox}
    FINAL VERIFICATION: \\\\
    Before giving your final answer, verify it:
    \begin{enumerate}
        \item Re-read the original question. List every requirement it contains.
        \item Check your answer against each requirement. Mark each as \checkmark\ satisfied or \xmark\ not satisfied.
        \item For any \xmark\ items: do you have enough information to fix it? If yes, fix it now. If no, note what's missing and give your best answer with what you have.
        \item Check formatting: does your answer match the requested format exactly? (number, string, list, etc.)
    \end{enumerate}
    Do NOT discard your work. Only correct specific errors you identified above.
\end{promptbox}

{\bf Trigger Condition}: injected once before the final answer

{\bf Design Rationale}: Even after good planning, diligent execution, and adaptive recovery, agents frequently produce answers that miss one or more requirements from the original question, especially when the question is multi-part or has specific formatting requirements. VERIFY acts as a safety net that catches these last-mile errors. The instruction "Do NOT discard your work" is critical: without it, agents sometimes over-correct by throwing away good work and starting over, which is wasteful and often produces worse results.

\section{Motivation: Backward Error Analysis for Agent Debugging}
\label{appendix:algorithm}

A central challenge in improving LLM-based agents is the opacity of their failure modes. When an agent produces an incorrect answer after a multi-step reasoning trajectory involving tool calls, web searches, and intermediate computations, it is rarely obvious where the reasoning went wrong. The final output is a single point of failure, but the causal chain leading to it may span dozens of steps. Without a principled method for localizing errors within this chain, practitioners are left with two unsatisfying options: (1) manually inspecting lengthy execution traces, which does not scale, or (2) blindly re-running the agent, which tends to reproduce the same errors.
We draw inspiration from back-propagation in neural network training, where the gradient of the loss is propagated backward through the computation graph to identify which parameters contributed most to the error. Analogously, we treat the agent's reasoning trace as a directed computation graph: a sequence of decisions, tool invocations, and intermediate conclusions. We propagate the error signal (an incorrect final answer) backward through this graph to identify the root cause: the earliest decision point where a different choice would have plausibly led to a correct outcome.

The backward analysis (PIVOT with HITL) revealed a consistent pattern: the most common failure mode is not missing information but ignored information. This diagnosis points to a structural problem: the agent's thinking budget is allocated in the wrong place. The agent loop is a simple cycle: \\
\hspace*{1cm}  {\bf call LLM → execute tool calls → when no more tool calls, extract plan $\rightarrow$ return}
\\\\
There are no intermediate steps between ``done gathering data'' and ``generate plan'' where the agent cross-references its collected evidence against constraints. More importantly, the same four failure archetypes recur systematically across tasks and models:

\begin{itemize}
    \item Absence of planning: agents dive into tool calls without structuring their approach, producing disorganized trajectories with redundant searches and missed dependencies.
    \item Ignored evidence: agents retrieve all necessary data but fail to cross-reference it during output assembly — the dominant failure mode at 44–47\% of cases.
    \item Retry loops: when a tool call fails, agents retry the same query rather than pivoting to a fundamentally different strategy.
    \item Silent constraint dropping: agents satisfy most requirements but silently drop one or two constraints, producing partially correct answers indistinguishable from fully correct ones.
\end{itemize}

This observation directly motivates the PIVOT framework: rather than diagnosing and repairing failures post-hoc, we embed the lessons from backward analysis as proactive guardrails that prevent the failures from arising in the first place. Each PIVOT stage maps to one of the archetypes above: PLAN addresses absence of planning, INSPECT catches ignored evidence mid-execution, EVOLVE breaks retry loops, and VERIFY eliminates silent constraint dropping at the final gate.

\begin{algorithm}[htb]
\caption{PIVOT: PLAN--INSPECT--EVOLVE--VERIFY Trajectory Refinement}
\label{alg:pivot}
\begin{algorithmic}[1]
\Require task $t$, pool size $M$, refinement steps $R$

\State $\mathcal{T}_0 \gets \{\texttt{\textsc{Plan}}(t)\}_{m=1}^{M}$
\For{each $\tau \in \mathcal{T}_0$}
    \State Execute $\tau \rightarrow \hat{\tau}$
    \State $\widehat{\mathcal{L}}(\tau) \gets \texttt{\textsc{Inspect}}(\hat{\tau}, \tau, t)$
\EndFor

\State $\tau^{(0)} \gets \operatorname*{arg\,min}_{\tau \in \mathcal{T}_0}^{\prec} \widehat{\mathcal{L}}(\tau)$
\State $\tau^\star \gets \tau^{(0)}$
\State $\widehat{\mathcal{L}}^\star \gets \widehat{\mathcal{L}}(\tau^{(0)})$

\For{$r = 0$ to $R-1$}
    \State Execute $\tau^{(r)} \rightarrow \hat{\tau}^{(r)}$
    \State $(\widehat{\mathcal{L}}^{(r)}, g^{(r)}) \gets \texttt{\textsc{Inspect}}(\hat{\tau}^{(r)}, \tau^{(r)}, t)$

    \If{$\mathrm{Success}(\hat{\tau}^{(r)}, t)$}
        \State \Return $\texttt{\textsc{Verify}}(\tau^{(r)}, t)$
    \EndIf

    \State $\tau' \gets \texttt{\textsc{Evolve}}(\hat{\tau}^{(r)}, \tau^{(r)}, g^{(r)})$

    \State Execute $\tau' \rightarrow \hat{\tau}'$
    \State $(\widehat{\mathcal{L}}', \_) \gets \texttt{\textsc{Inspect}}(\hat{\tau}', \tau', t)$

    \If{$\widehat{\mathcal{L}}' \prec \widehat{\mathcal{L}}^{(r)}$}
        \State $\tau^{(r+1)} \gets \tau'$
        \State $\widehat{\mathcal{L}}^{(r+1)} \gets \widehat{\mathcal{L}}'$
    \Else
        \State $\tau^{(r+1)} \gets \tau^{(r)}$
        \State $\widehat{\mathcal{L}}^{(r+1)} \gets \widehat{\mathcal{L}}^{(r)}$
    \EndIf
\EndFor
\State $\tau^\star \gets \tau^{(R)}$
\State \Return $\texttt{\textsc{Verify}}(\tau^\star, t)$
\end{algorithmic}
\end{algorithm}

\section{Sampled PIVOT Trajectories (self-judged)}

\subsection{Example 1 — id\_16: Hangzhou$\rightarrow$ Shaoxing day trip (0.375 $\rightarrow$  1.000)}

{\bf Query}: Solo traveler, earliest train, cheapest Home Inn, must eat near Bazhi Bridge at a restaurant with waiting area service.\\
{\bf What the baseline got wrong}: Scheduled an attraction visit before opening hours (Lu Xun's Hometown at 08:23, opens 08:30), missed the ``waiting area service'' restaurant tag requirement near Bazhi Bridge, and had an incomplete Day 2 itinerary.\\ 
{\bf What PIVOT audit caught}: The <thinking> block systematically listed all constraints as a checklist, flagged the opening-hours violation (``Lu Xun's Hometown opens at 08:30, not 08:23!''), identified the missing restaurant tag constraint, and noted the incomplete Day 2. It then rebuilt the plan from scratch, fixing all three issues to achieve a perfect 1.000 composite score.

\subsection{Example 2 — id\_62: Harbin $\rightarrow$ Beijing 5-day trip (0.125 $\rightarrow$  0.875)}

{\bf Query}: 3 travelers, direct Airbus flight (cheapest), highest-rated 3-star hotel, 2 rooms, must eat at ``Pebbles Courtyard Mexican Restaurant,'' restaurant with private room near Beihai Park.\\
{\bf What the baseline got wrong}: Selected a Boeing flight instead of Airbus (violating the aircraft manufacturer constraint), missed the must-eat restaurant entirely, scheduled a meal outside restaurant operating hours, and had cost calculation errors across the budget summary.\\
{\bf What PIVOT audit caught}: The <thinking> block enumerated all available flights and marked each Boeing option with \xmark, identifying CZ6217 (Airbus 320, ¥520) as the correct cheapest Airbus direct flight. It flagged the restaurant operating hours violation and the missing must-eat restaurant. The rebuilt plan corrected the flight, included Pebbles Courtyard, and fixed the budget — recovering 6 of 8 commonsense dimensions.

\subsection{Example 3 — id\_58: Shenzhen$\rightarrow$ Beijing 4-day trip (0.250 $\rightarrow$  0.938)}

{\bf Query}: 4 travelers, return flight arriving 11:00–15:00, 3-star hotel with gym, birthday dinner near Sanlitun Taikoo Li, budget meal near Tiananmen.\\
{\bf What the baseline got wrong}: Selected return flight ZH1393 arriving at 10:10 AM (before the 11:00 AM requirement), reused bluefrog restaurant on two different days, and routed through Shichahai without allocating actual visit time.\\
{\bf What PIVOT audit caught}: Flagged 4 critical failures — return flight time violation (``\xmark\ FAILED — arrives at 10:10 AM, BEFORE 11:00''), restaurant diversity violation (``\xmark\ — I used bluefrog twice''), incomplete attraction coverage ("\xmark\  — Shichahai never actually visited"). The rebuilt plan switched to flight DZ6210 (arrives 11:55 AM), replaced the duplicate restaurant, and properly scheduled Shichahai.

\subsection{Example 4 — id\_19: Ningbo$\rightarrow$ Suzhou 2-day trip (0.312 $\rightarrow$  0.938)}

{\bf Query}: 4 travelers, most affordable direct train, hotel ¥330–360/night, 2 rooms, highest-rated restaurant near Suzhou Museum, restaurant with outdoor seating near Suzhou Museum. \\
{\bf What the baseline got wrong}: Selected a hotel (Four Points by Sheraton Kunshan) outside the ¥330–360 price range, had no activity diversity (insufficient attractions), and violated itinerary structure rules. \\
{\bf What PIVOT audit caught}: Identified the hotel price constraint violation and the missing outdoor-seating restaurant tag. The rebuilt plan selected a correctly-priced hotel, added diverse attractions, and matched both restaurant requirements near Suzhou Museum.

\subsection{Example 5 — id\_32: Harbin$\rightarrow$ Beijing 3-day trip (0.188 $\rightarrow$  0.812)}

{\bf Query}: Solo traveler, first-class train both ways, 3-star hotel with SPA, restaurant with waiting area near Shichahai, cheapest restaurant near Tsinghua University. \\
{\bf What the baseline got wrong}: Failed to find the cheapest restaurant near Tsinghua University (used wrong entity), had route consistency errors (teleportation between locations), violated business hours, and miscalculated costs.\\
{\bf What PIVOT audit caught}: Flagged ``\xmark\ CRITICAL FAILURE — Meal 2: Near Tsinghua University, MUST pick LOWEST average cost per person'' and identified the wrong restaurant selection. The rebuilt plan queried the correct cheapest option (Tsinghua University Nanyuan Restaurant) and fixed route continuity.

\subsection{Example 6 — id\_36: Wuhan$\rightarrow$ Nanjing 3-day trip (0.188 $\rightarrow$  0.812)}

{\bf Query}: 3 travelers, earliest direct train, 4-star hotel with TV screen mirroring, 2 rooms, budget $\geq$ \textyen 5,500, restaurant with online queue near Nanjing Eye, must eat at ``New Bailu.''\\
{\bf What the baseline got wrong}: Selected a hotel without TV screen mirroring capability (Atour Hotel instead of one with the required service), had unreasonable visit durations, and violated business hours.\\
{\bf What PIVOT audit caught}: Flagged "\xmark\ CRITICAL FAILURE — Accommodation: TV screen mirroring" and "I chose the WRONG hotel." The rebuilt plan selected a hotel with the correct amenity and restructured the daily schedule with proper durations.

\section{Sampled PIVOT Trajectories (oracle)}

\subsection{Example 1— id\_49: Shanghai$\rightarrow$ Enshi 4-day trip (0.312 $\rightarrow$  0.875)}

{\bf What the baseline got wrong}: Completely omitted ``Canyon Flavor Restaurant'' — a must-eat restaurant explicitly requested by the user — resulting in 0\% personalized score.\\
{\bf What backward analysis diagnosed}: The agent queried query\_restaurant\_details(``Canyon Flavor Restaurant'') and got ``not found,'' then gave up entirely instead of trying alternative name variations (``Grand Canyon Flavor Restaurant'') or coordinate-based searches. The backward analysis identified this as an INFORMATION error at the tool-retry level. The correction pass used alternative search strategies to locate and include the restaurant.

\subsection{Example 2 — id\_18: Hangzhou$\rightarrow$ Shaoxing 2-day trip (0.188 $\rightarrow$  0.875)}

{\bf Query}: Train departing 6:00–7:00 AM, cheapest hotel with swimming pool, must visit Keyan Scenic Area and free attractions, must eat at a specific local restaurant.\\
{\bf What the baseline got wrong}: 5 cascading violations — train price mismatch (¥16 vs actual ¥520), Keyan visited at 17:35 (closes 17:00), only 50 minutes at Keyan (requires 2.5–4 hours), unreasonable transfer gaps, and missing Shaoxing Museum.\\
{\bf What backward analysis diagnosed}: Root cause type INFORMATION. The agent's train query returned truncated results with only one option (G7711 at 16:30). The agent fabricated train G7713 at 06:40 which didn't exist in the results, causing a late arrival that compressed the entire Day 1 schedule into an impossible timeline. The correction pass (8 new tool calls) re-queried trains, restructured both days, and added the missing museum.

\subsection{Example 3 — id\_42: Urumqi$\rightarrow$ Shanghai 4-day trip (0.125$ \rightarrow$  0.750)}

{\bf Query}: Return flight arriving 14:00–18:00, 4-star hotel, must visit The Bund and Yu Garden, restaurant with birthday set menu.\\
{\bf What the baseline got wrong}: Only generated 3 days instead of 4, selected return flight arriving at 21:40 (outside the 14:00–18:00 window), and missed the birthday restaurant requirement.\\
{\bf What backward analysis diagnosed}: Root cause type INFORMATION. The agent retrieved the return flight but failed to validate the arrival time against the user's explicit 14:00–18:00 constraint. It accepted the first result without checking alternatives. The correction pass (3 new tool calls) found a compliant flight and rebuilt the 4-day itinerary.

\subsection{Example 4 — id\_82: Wuhan$\rightarrow$ Nanjing 6-day trip (0.125 $\rightarrow$  0.750)}

{\bf Query}: 6-day trip (Nov 12–17), train both ways, 3-star hotel, must visit historical sites, restaurant near Confucius Temple.\\
{\bf What the baseline got wrong}: Malformed last-day structure, insufficient attractions on final day, business hours violations, and timing gaps throughout.\\
{\bf What PIVOT with HITL analysis diagnosed}: Root cause type PLANNING. The agent correctly counted 5 nights but failed to convert this into 6 calendar days of itinerary. The Nov 12–17 span requires 6 daily schedules, but the agent generated fewer, leaving the final day incomplete. The correction pass (10 new tool calls) rebuilt the full 6-day structure with proper daily closure.

\subsection{Example 5 — id\_77: Ningbo$\rightarrow$ Zhengzhou 5-day trip (0.062 $\rightarrow$ 0.688)}

{\bf Query}: Hotel ¥250–280/night, must visit Yellow River scenic area, earliest train, restaurant near Erqi Square.\\
{\bf What the baseline got wrong}: Selected a hotel outside the ¥250–280 price range, had invalid transportation segments, missing meals, attractions visited outside operating hours, and incorrect visit durations.\\
{\bf What PIVOT with HITL analysis diagnosed}: Root cause type REASONING. The agent did not enforce the user's explicit numerical hotel price constraint — treating ``between 250 and 280 yuan per night'' as a guideline rather than a hard boundary. The correction pass (23 new tool calls) re-queried hotels with strict price filtering and rebuilt the entire itinerary from scratch.

\subsection{Example 6 — id\_117: Chongqing$\rightarrow$ Zhengzhou 7-day trip (0.312 $\rightarrow$ 0.875)}

{\bf Query}: Highest-rated Home Inn, must visit Shaolin Temple and Yellow River, restaurant with private room near Erqi Square, budget-friendly meal near Zhengzhou East Station.\\
{\bf What the baseline got wrong}: Did not book the highest-rated Home Inn (4.9$\star$) despite retrieving it, missed required attractions, and failed personalized restaurant constraints.\\
{\bf What PIVOT with HITL analysis diagnosed}: Root cause type REASONING $\rightarrow$  INFORMATION. The agent was ``data-sufficient but logic-deficient'' - it gathered all necessary information through tool calls but did not apply systematic filtering when assembling the plan. It retrieved the 4.9$\star$ hotel but selected a lower-rated one; it found the required attractions but didn't include them all. The correction pass (17 new tool calls) applied proper selection logic to the already-gathered data.

\section{Experiment Implementation Details}
\label{appendix:details}

\subsection{Experiments Compute Resource}
All the experiments related to the LLM inference are conducted by Amazon Bedrock platform. For the data analysis part, we used the Amazon EC2 g5.48xlarge instance, with 8 NVIDIA A10G GPUs (24 GB memory each). All experiments are run through the AWS Bedrock Converse API with temperature $0.0$ for deterministic decoding. When extended thinking is enabled (Claude 4.5 Haiku / Sonnet / Opus and Claude 4.6 Opus), temperature is set to $1.0$ as required by the Claude API. Qwen3-235B-A22B does not support extended thinking and is always run at temperature $0.0$.

\subsection{Definition of Tools in GAIA Evaluation}
\label{app:gaia_tools}

GAIA tasks are solved by an agent that interacts with four tools through the Bedrock Converse \texttt{toolConfig} interface. The same tool set is used for every method compared in Section~\ref{sec:experiment}, so differences in task accuracy reflect reflection strategy rather than tool affordance. Each tool is exposed to the model with a JSON schema specifying its name, a one-sentence description, and a typed input object; the agent selects tools via the native tool-use protocol.

\textbf{Tool Catalog.}
\begin{itemize}[leftmargin=1.4em,itemsep=1pt]
    \item \texttt{web\_search(query: str)}. Executes a Google query and returns the top result snippets. Falls back to a DuckDuckGo lite scrape if Google returns no parseable snippets. Output is limited to the first five results.
    \item \texttt{web\_fetch(url: str)}. Retrieves a URL with a 20s timeout, strips navigation, scripts, and style tags from HTML, and returns the visible text. Content is truncated to 15{,}000 characters.
    \item \texttt{read\_file(filename: str)}. Reads an attachment associated with the task. PDFs, Word (\texttt{.docx}), Excel (\texttt{.xlsx}), PowerPoint (\texttt{.pptx}), CSV, JSON, Python, and plain-text files are returned as text. Images (\texttt{.png}, \texttt{.jpg}) are returned as a vision-model caption. Audio (\texttt{.mp3}) is returned as an ASR transcription.
    \item \texttt{python\_exec(code: str)}. Executes Python code in an isolated subprocess with a 30-second wall-clock limit; standard output (and any stderr) is returned, truncated to 10{,}000 characters. Pre-installed libraries include \texttt{pandas}, \texttt{numpy}, \texttt{json}, \texttt{math}, \texttt{re}, \texttt{collections}, and \texttt{itertools}.
\end{itemize}

\paragraph{Scope.} This tool set is deliberately minimal. It does not include a browser, a multi-hop retriever, or any specialist tool (e.g., PubMed search, symbolic solver). Every method, ours and the baselines, operates under the same ceiling on what information can be recovered from the environment. Accuracy differences attributed to PIVOT therefore reflect better use of the available tools, not additional affordances.

\end{document}